\documentclass[10pt,twocolumn,letterpaper]{article}

\usepackage{iccv}
\usepackage{times}
\usepackage{epsfig}
\usepackage{graphicx}
\usepackage{amsmath}
\usepackage{amssymb}
\usepackage{multirow}
\usepackage{dutchcal}
\usepackage{subfigure}

\usepackage[pagebackref=true,breaklinks=true,letterpaper=true,colorlinks,bookmarks=false]{hyperref}

\iccvfinalcopy 

\def\iccvPaperID{3352} 

\ificcvfinal\fi

\begin{document}

\title{Skeleton Cloud Colorization for Unsupervised 3D Action \\Representation Learning}

\author{Siyuan Yang$^{1}$ ~\quad \quad Jun Liu$^{2}$\thanks{Corresponding author.} \quad\quad ~Shijian Lu$^1$ \quad\quad ~Meng Hwa Er$^1$  \quad\quad~ Alex C. Kot$^1$ \quad\quad
\\
$^1$Nanyang Technological University ~~~~
$^2$Singapore University of Technology and Design \\
{\tt\small siyuan005@e.ntu.edu.sg \quad ~~jun\_liu@sutd.edu.sg ~~\quad \{Shijian.Lu, emher, eackot\}@ntu.edu.sg}
}


\maketitle
\ificcvfinal\thispagestyle{empty}\fi

\begin{abstract}
Skeleton-based human action recognition has attracted increasing attention in recent years. However, most of the existing works focus on supervised learning which requiring a large number of annotated action sequences 
that are often expensive to collect. We investigate unsupervised representation learning for skeleton action recognition, and design a novel skeleton cloud colorization technique that is capable of learning skeleton representations from unlabeled skeleton sequence data. Specifically, we represent a skeleton action sequence as a 3D skeleton cloud and colorize each point in the cloud according to its temporal and spatial orders in the original (unannotated) skeleton sequence. Leveraging the colorized skeleton point cloud, we design an auto-encoder framework that can learn spatial-temporal features from the artificial color labels of skeleton joints effectively. 
We evaluate our skeleton cloud colorization approach with action classifiers trained under different configurations, including unsupervised, semi-supervised and fully-supervised settings.
Extensive experiments on NTU RGB+D and NW-UCLA datasets show that the proposed method outperforms existing unsupervised and semi-supervised 3D action recognition methods by large margins, and it achieves competitive performance in supervised 3D action recognition as well.
\end{abstract}

\begin{figure*}[t]
\begin{center}
\includegraphics[trim=1.2cm 1.2cm 1.2cm 0.7cm,clip,width=0.9\textwidth]{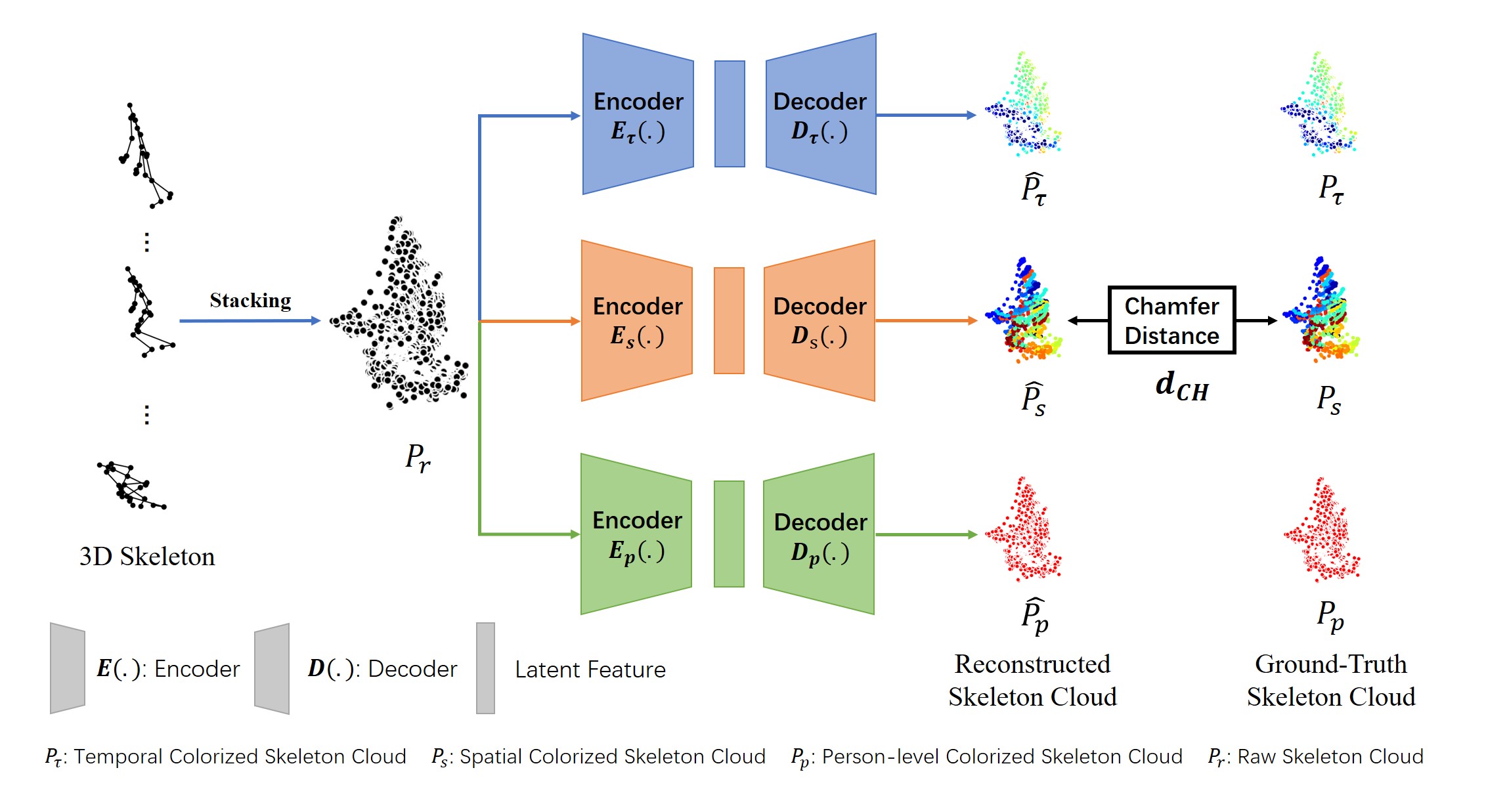}
 \vspace{-0.3cm}
\end{center}
      \caption{
The pipeline of our proposed unsupervised representation learning method, using a novel skeleton cloud colorization scheme. Given a 3D skeleton sequence, we first stack it into a raw skeleton cloud $P_{r}$ and then colorize it into 3 skeleton clouds $P_{\tau}$, $P_{s}$, and $P_{p}$ (construction details shown in Figs. \ref{fig:color_pipeline} and \ref{fig:person pipeline}) according to spatial, temporal, and person-level information, respectively. With the three colorized clouds as self-supervision signal, three encoder-decoders (with the same structure but no weight sharing) learn discriminative skeleton representative features.
(The encoder and decode details are provided in supplementary material.)
  }
  \label{fig:framework}
\end{figure*}

\section{Introduction}
Human action recognition is a fast developing area due to its wide applications in human-computer interaction, video surveillance, game control, etc. According to the types of input data, human action recognition can be grouped into different categories such as RGB-based \cite{carreira2017quo, simonyan2014two, tran2015learning, wang2018temporal, xie2018rethinking, yang2020collaborative, zhu2017tornado, zhu2018yotube}, depth-based \cite{ oreifej2013hon4d, rahmani2014hopc, wang20203dv} and 3D skeleton-based \cite{yujuniccv2021, ke2017new, li2021elsenet, Li_2021_CVPR, liu2017skeleton, liu2016spatio, Shi_2019_CVPR_twostream}, etc. Among these types of inputs, 3D skeleton data, which represents a human body by the locations of keypoints in the 3D space, has attracted increasing attention in recent years. Compared with RGB videos or depth data, 3D skeleton data encodes high-level representations of human behaviors and it is generally lightweight and robust to the variations in appearances, surrounding distractions, viewpoint changes, etc. Additionally, skeleton sequences can be easily captured by depth sensors and thus a large number of supervised methods have been designed to learn spatio-temporal representations for skeleton based action recognition.

Deep neural networks have been widely studied to model the spatio-temporal representation of skeleton sequences under supervised scenarios \cite{ke2017new, liu2017skeleton, liu2016spatio, Shi_2019_CVPR_twostream}. For example, Recurrent Neural Network (RNN) has been explored for modelling skeleton actions since it can capture temporal relations well \cite{du2015hierarchical, liu2017skeleton,liu2016spatio,Shahroudy_2016_CVPR, zhang2017view}. Convolutional Neural Networks (CNNs) have also been explored to build skeleton-based recognition frameworks by converting joint coordinates to 2D maps \cite{du2015skeleton, ke2017new, li2017skeleton, wang2017scene}. In addition, Graph Convolution Networks (GCNs) have attracted increasing attention due to their outstanding performance \cite{peng2020learning, Shi_2019_CVPR_twostream, yan2018spatial}. However, all these methods are supervised which require a large number of labelled training samples that are often costly to collect. How to learn effective feature representations with minimal annotations thus becomes critically important. To the best of our knowledge, only a few works \cite{kundu2019unsupervised, li2020sparse, lin2020ms2l, su2020predict, zheng2018unsupervised} explored representation learning from unlabelled skeleton data for the task of action recognition, where the major approach is to reconstruct skeleton sequences from the encoded features via certain encoder-decoder structures.
Unsupervised skeleton-based action representation learning remains a great challenge.

In this work, we propose to represent skeleton sequences as a 3D skeleton cloud, and design an unsupervised representation learning scheme that learns features from spatial and temporal color labels. We treat a skeletal sequence as a spatial-temporal skeleton cloud by stacking the skeleton data of all frames together, and colorize each point in the cloud according to its temporal and spatial orders in the original skeleton sequence. Specifically, we learn spatial-temporal features from the corresponding joints' colors by leveraging a point-cloud based auto-encoder framework as shown in Fig.~\ref{fig:framework}. 
By repainting the whole skeleton cloud, our network can achieve unsupervised skeleton representation learning successfully by learning both spatial and temporal information from unlabeled skeleton  sequences.

The contributions of this paper are threefold. \textit{First}, we formulate unsupervised action representation learning as 3D skeleton cloud repainting problem, where each skeleton sequence is treated as a skeleton cloud and can be directly processed with a point cloud auto-encoder framework. 
\textit{Second}, we propose a novel skeleton cloud colorization scheme that colorizes each point in the skeleton cloud based on its temporal and spatial orders in the skeleton sequence. The color labels `fabricate’ self-supervision signals which boost unsupervised skeleton action representation learning significantly. \textit{Third}, extensive experiments show that our method outperforms state-of-the-art unsupervised and semi-supervised skeleton action recognition methods by large margins, and its performance is also on par with supervised skeleton-based action recognition methods.

To the best of our knowledge, this is the first work that converts the unsupervised skeleton representation learning problem into a novel skeleton cloud repainting task.

\section{Related work}
{\bf Skeleton-based action recognition.} Skeleton-based action recognition has attracted increasing interest recently. Unlike traditional methods that design hand-craft features \cite{hussein2013human, vemulapalli2014human, wang2013learning,xia2012view}, deep-learning based methods employ Recurrent Neural Networks (RNNs), Convolutional Neural Networks (CNNs), and Graph Convolution Networks (GCNs) to learn skeleton-sequence representation directly. Specifically, RNNs have been widely used to model temporal dependencies and capture the motion features for skeleton-based action recognition. For example, \cite{du2015hierarchical} uses a hierarchical RNN model to represent human body structures and temporal dynamics of the body joints. \cite{liu2017skeleton,liu2016spatio} proposes a 2D Spatio-Temporal LSTM framework to employ the hidden sources of action related information over both spatial and temporal domains concurrently. \cite{zhang2017view} adds a view-adaptation scheme to the LSTM to regulate the observation viewpoints.

CNN-based methods \cite{du2015skeleton, ke2017new, li2017skeleton, wang2017scene} have also been proposed for skeleton action recognition. They usually transform the skeleton sequences to skeleton maps of same target size and then use CNNs to learn the spatial and temporal dynamics. For example, \cite{du2015skeleton, li2017skeleton} transform a skeleton sequence to an image by treating the joint coordinate (x,y,z) as the R, G, and B channels of a pixel. \cite{ke2017new} transforms the 3D skeleton data to three skeleton clips for robust action feature learning. \cite{wang2017scene} presents a “scene flow to action map” representation for action recognition with CNNs.

Inspired by the observation that human 3D-skeleton is naturally a topological graph, Graph Convolutional Networks (GCN) have attracted increasing attention in skeleton-based action recognition. For example, \cite{yan2018spatial} presents a spatial-temporal GCN to learn both spatial and temporal patterns from skeleton data. \cite{Shi_2019_CVPR_twostream} uses a non-local method with the spatial-temporal GCN to boost performance. \cite{Shi_2019_CVPR_dgnn} uses bone features for skeleton-based action recognition. \cite{peng2020learning} recognizes actions by searching for different graphs at different layers via neural architecture search.

Though the aforementioned methods achieve very impressive performance, they are all supervised requiring large amount of labelled data which is prohibitively time-consuming to collect. In this work, we study unsupervised representation learning in skeleton-based action recognition which mitigates the data labelling constraint greatly.

{\bf Unsupervised representation learning for action recognition.} Unsupervised action recognition aims to learn effective feature representations by predicting future frames of input sequences or by re-generating the sequences. Most existing methods focus on RGB videos or RGB-D videos. For example, \cite{srivastava2015unsupervised} uses a LSTM based Encoder-Decoder architecture to learn video representations \cite{luo2017unsupervised} uses a RNN based encoder-decoder framework to predict the sequences of flows computed with RGB-D modalities. \cite{li2018unsupervised} uses unlabeled video to learn view-invariant video representations.

Unsupervised skeleton-based action recognition was largely neglected though a few works attempt to address this challenging task very recently. For example, \cite{zheng2018unsupervised} presents a GAN encoder-decoder to re-generate masked input sequences. \cite{kundu2019unsupervised} adopts a hierarchical fusion approach to improve human motion generation. \cite{su2020predict} presents a decoder-weakening strategy to drive the encoder to learn discriminative action features.

The aforementioned methods all process skeleton sequences frame by frame and extract temporal features from ordered sequences. We instead treat a skeleton sequence as a novel colored skeleton cloud by stacking human joints of each frame together. We design a novel skeleton colorization scheme and leverage the color information for unsupervised spatial-temporal representation learning.


\section{Method}

In this section, we present our skeleton cloud colorization representation learning method that converts the skeleton sequence to a skeleton cloud, and colorizes each point in the cloud by its spatial-temporal properties. In particular, we present how to construct the skeleton cloud in Section \ref{Data_process} and describe the colorization step in Section \ref{Skeleton Coloring}.
Repainting pipeline and training details are discribed in Section \ref{Ratio of mask} and Section \ref{training}, respectively.


\subsection{Data Processing}\label{Data_process}
Given a skeleton sequence $S$ under the global coordinate system, the $j^{th}$ skeleton joint in the $t^{th}$ frame is denoted as $v_{t, j} = [x_{t, j}, y_{t, j}, z_{t, j}]$, $t\in (1, \cdots, T)$, $j\in (1, \cdots, J)$, where $T$ and $J$ denote the number of frames and body joints, respectively. Generally, skeleton data is defined as a sequence and the set of joints in the $t^{th}$ frame are denoted as $V_{t} = \left\{v_{t,j}|j = 1, ..., J  \right\}$. We propose to treat all the joints in a skeleton sequence as a whole by stacking all frames' data together, and Fig.~\ref{fig:framework} illustrates the stacking framework. We name the stacked data as skeleton cloud and denote it by $P_{r} = \left\{v_{t, j} = [x_{t, j}, y_{t, j}, z_{t, j}]|t = 1, ..., T; j = 1, ..., J \right\}$. The obtained 3D skeleton cloud therefore consists of $N = T\times J$ 3D points in total. We use $P_{r}$ to denote raw skeleton cloud so as to differentiate it from the colorized clouds to be described later.

\begin{figure}[t]
\begin{center}
\includegraphics[width=0.26\textwidth]{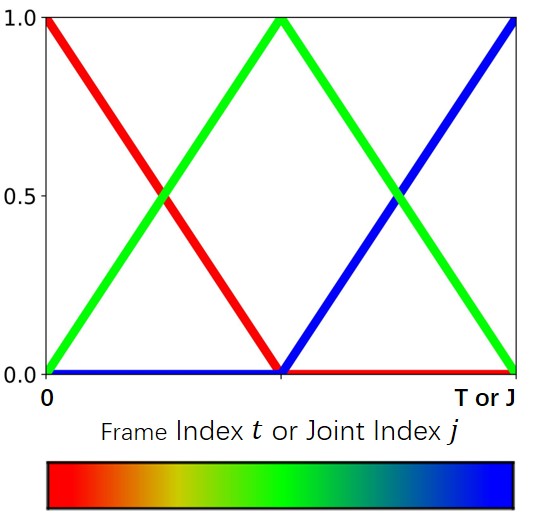}
\vspace{-0.39cm}
\end{center}
  \caption{Illustration of the colorization scheme for temporal index $t$ and spatial index $j$.
  Top: Definition of each color channel(RGB) when varying $t$ or $j$ (where $t \in [1, T]; j \in [1, J]$). 
  Bottom: Corresponding color of the temporal index $t$ and spatial index $j$.
  With the increase of the point's temporal/spatial order index, the corresponding color changes from red to green to blue. (Best viewed in color) 
  }
\label{fig:colorize}
\end{figure}

\subsection{Skeleton Cloud Colorization}\label{Skeleton Coloring}
Points within our skeleton cloud are positioned with 3D spatial coordinates (x, y, z) which is similar to normal point cloud that consists of unordered points. The spatial relation and temporal dependency of skeleton cloud points are crucial in skeleton-based action recognition, but they are largely neglected in the aforementioned raw skeleton cloud data. We propose an innovative skeleton cloud colorization method to exploit the spatial relation and temporal dependency of skeleton cloud points for unsupervised skeleton-based action representation learning.

\begin{figure*}[t]
\begin{center}
\includegraphics[trim=0.5cm 0.5cm 2cm 0.1cm,clip,width=0.87\textwidth]{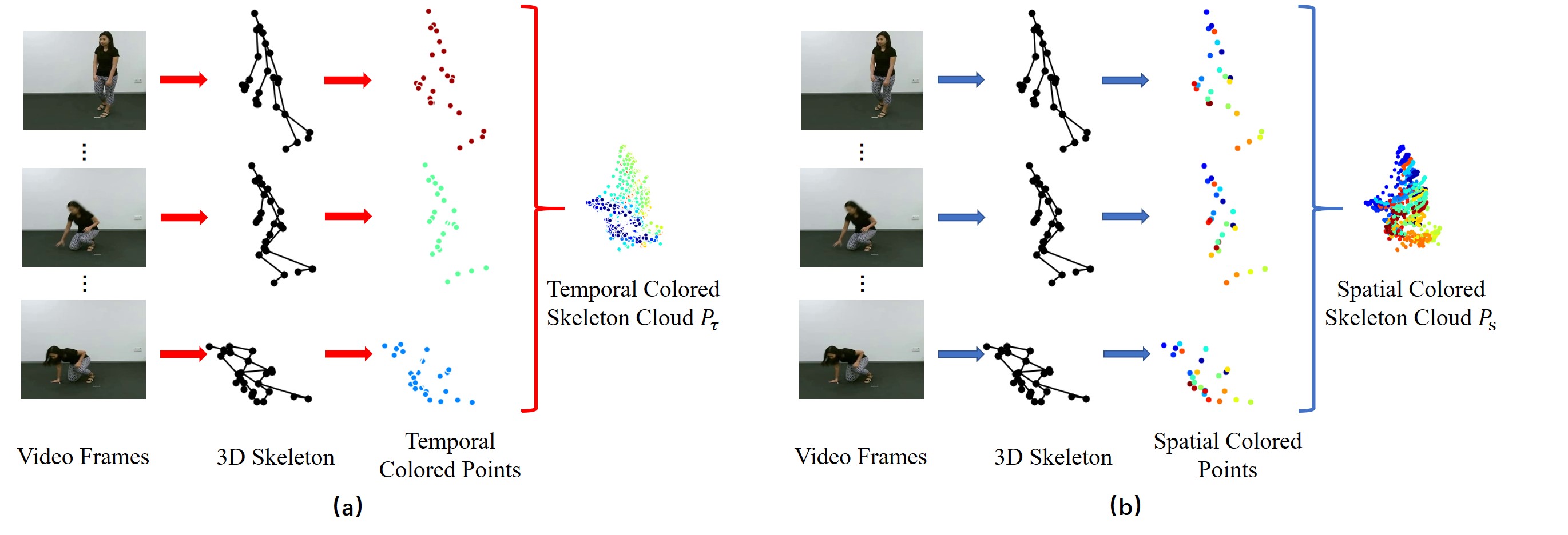}
 \vspace{-0.399cm}
\end{center}
\caption{
The pipelines of temporal colorization and spatial colorization. (a) Given a skeleton sequence, the temporal colorization colorizes points based on the relative temporal order $t$ ($t \in [1, T]$) in the sequential data. (b) The spatial colorization colorizes points based on the index of joints $j$ ($j \in [1, J]$). (Best viewed in color)  
}
\label{fig:color_pipeline}
\end{figure*}

\textbf{Temporal Colorization:} Temporal information is critical in action recognition. To assign each point in the skeleton cloud a temporal feature, we colorize the skeleton cloud points according to their relative time order (from $1$ to $T$) in the original skeleton sequence. Different colorization schemes have been reported and here we adopt the colorization scheme that uses 3 RGB channels \cite{choutas2018potion} as illustrated in Fig.~\ref{fig:colorize}. 
The generated color is actually relatively linear under this color scheme.
Hence, points from adjacent frames are assigned with similar color under this distribution, which facilitates the learning of temporal order information.
The value distributions of R, G, B channels can be formulated as follows:
\begin{equation}
\small
r^{\tau}_{t, j} = \left\{
\begin{aligned}
-2\times (t/T) + 1 &, \;\mbox{if $t <= T/2$}\\
0  &, \;\mbox{if $t > T/2$}\\
\end{aligned}
\right.
\end{equation}

\begin{equation}
\small
g^{\tau}_{t, j} = \left\{
\begin{aligned}
2\times (t/T) & ,\;\mbox{if $t <= T/2$}\\
-2\times (t/T) + 2  &, \;\mbox{if $t > T/2$}\\
\end{aligned}
\right.
\end{equation}

\begin{equation}
\small
b^{\tau}_{t, j} = \left\{
\begin{aligned}
0 & ,\;\mbox{if $t <= T/2$}\\
2\times (t/T) - 1  & ,\;\mbox{if $t > T/2$}\\
\end{aligned}
\right.
\end{equation}

With this colorizing scheme, we can assign different colors to points from different frames based on the frame index $t$ as illustrated in Fig.~\ref{fig:color_pipeline}(a). More specifically, with this temporal-index based colorization scheme, each point will have a 3-channels features that can be visualized with red, green and blue channels (RGB channels) to represent its temporal information. Together with the original 3D coordinate information, the temporally colorized skeleton cloud can be denoted by $P_{\tau} = \{v^{\tau}_{t, j} = [x_{t, j}, y_{t, j}, z_{t, j}, r^{\tau}_{t, j}, g^{\tau}_{t, j}, b^{\tau}_{t, j}]|t = 1, ..., T; j = 1, ..., J \}$.

\textbf{Spatial Colorization:} Besides temporal information, spatial information is also very important for action recognition. We employ similar colorization scheme to colorize spatial information as illustrated in Fig.~\ref{fig:colorize}. The scheme assigns different colors to different points according to their spatial orders $j \in [1, J]$ ($J$ is the total number of joints in the skeleton cloud of a person), as shown in Figure \ref{fig:color_pipeline}(b). The distribution of the values of R, G, B channels can be calculated as follows:
\begin{equation}
\small
r^{s}_{t, j} = \left\{
\begin{aligned}
-2\times (j/J) + 1 & ,\;\mbox{if $j <= J/2$}\\
0  & ,\;\mbox{if $j > J/2$}\\
\end{aligned}
\right.
\end{equation}

\begin{equation}
\small
g^{s}_{t, j} = \left\{
\begin{aligned}
2\times (j/J) & ,\;\mbox{if $j <= J/2$}\\
-2\times (j/J) + 2  & ,\;\mbox{if $j > J/2$}\\
\end{aligned}
\right.
\end{equation}

\begin{equation}
\small
b^{s}_{t, j} = \left\{
\begin{aligned}
0 & ,\;\mbox{if $j <= J/2$}\\
2\times (j/J) - 1  & ,\;\mbox{if $j > J/2$}\\
\end{aligned}
\right.
\end{equation}

We denote the spatially colorized skeleton cloud as $P_{s} = \{ v^{s}_{t, j} = [x_{t, j}, y_{t, j}, z_{t, j}, r^{s}_{t, j}, g^{s}_{t, j}, b^{s}_{t, j}]|t = 1, ..., T; j = 1, ..., J \}$. With the increase of the spatial order index of the joint in the skeleton, points will be assigned with different colors that change from red to blue and to green gradually.

\begin{figure}[t]
\begin{center}
\includegraphics[trim=4cm 1.5cm 1.5cm 2cm,clip,width=0.35\textwidth]{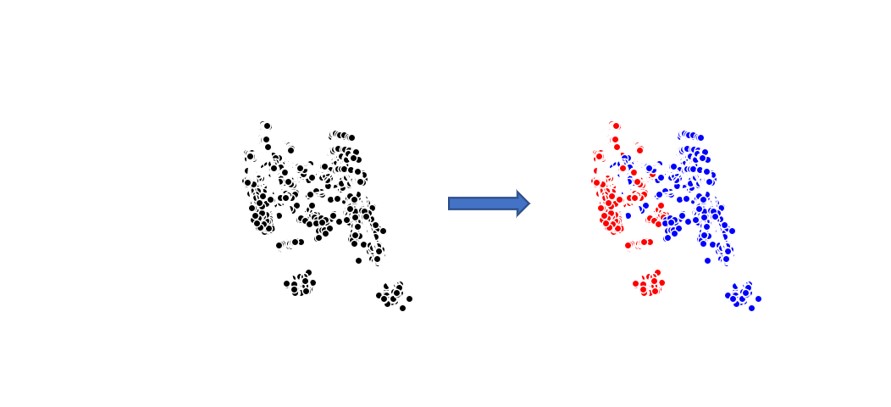}
\end{center}
 \vspace{-0.390cm}
  \caption{Person-level colorization. The first person's points will be assigned the red color, and person two will colorize to blue. 
  }
 
\label{fig:person pipeline}
\end{figure}

\textbf{Person-level Colorization:} Human actions contain rich person interaction information as in NTU RGB+D \cite{Shahroudy_2016_CVPR} which is important to the skeleton action recognition. We therefore propose a person-level colorization scheme for action recognition.

We focus on the scenarios that human interactions involve two persons only, and apply different colors to the points of different persons. Specifically, we encode the first person's joints with red color and the second person's joints with blue color as illustrated in Fig.~\ref{fig:person pipeline}. The person-level colored clouds can thus be denoted by $P_{p} = \{v^{p}_{t, j, n} = [x_{t, j, n}, y_{t, j, n}, z_{t, j, n}, 1, 0, 0]| t = 1, ..., T; j = 1, ..., J ; n = 1 \} \cup \{v^{p}_{t, j, n} = [x_{t, j, n}, y_{t, j, n}, z_{t, j, n}, 0, 0, 1]| t = 1, ..., T; j = 1, ..., J; n = 2 \} $ ,where $n = 1$ and $n = 2$ mean that the points belong to the first and the second persons, respectively. 

Given a raw skeleton cloud, the three colorization schemes thus construct three colorized skeleton clouds $P_{\tau}$, $P_{s}$ and $P_{p}$ that capture temporal dependency, spatial relations and human interaction information, respectively.

\subsection{Repainting Pipeline}\label{Ratio of mask}
Inspired by the success of self-supervised learning, our goal is to extract the temporal, spatial, and interactive information by learning to repaint the raw skeleton cloud $P_{r}$ in self-supervised manner. As illustrated in Fig.~\ref{fig:framework}, we use colorized skeleton clouds (temporal-level $P_{\tau}$, spatial-level $P_{s}$, and person-level $P_{p}$) as three kinds of self-supervision signals, respectively. The framework consists of an encoder $E(.)$ and a decoder $D(.)$. Since we have three colorization schemes, we have three pairs of encoders ($E_{\tau}(.)$, $E_{s}(.)$, and $E_{p}(.)$) and decoders ($D_{\tau}(.)$, $D_{s}(.)$, and $D_{p}(.)$). Below we use the temporal colorization stream as the example to explain the model architecture and the training process.

{\bf Model Architecture:} As mentioned in Section \ref{Skeleton Coloring}, the obtained skeleton cloud format is similar to that of the normal point cloud. We therefore adopt DGCNN \cite{wang2019dynamic} (designed for point cloud classification and segmentation) as the backbone of our framework and use the modules before the fully-connected (FC) layers to build our encoder \footnote{Detailed structure of encoder and decoder can be found in supplementary.\label{footnote}}.

In addition, we adopt the decoder \textsuperscript{\ref{footnote}} of FoldingNet \cite{yang2018foldingnet} as the decoder of our network architecture. Since the input and output of FoldingNet are all $N\times3$ matrices with 3D positions $(x, y, z)$ only, we enlarge the feature dimension to 6 to repaint both the position and color information. Assuming that the input is the raw point set $P_{r}$ and the obtained repainted point set is $\widehat{P_{\tau}} = D_{\tau}(E_{\tau}(P_{r}))$, the repainting error between the ground truth temporal colorization $P_{\tau}$ and the repainted $\widehat{P_{\tau}}$ is computed by using the Chamfer distance:
\begin{equation}
\small
    d_{CH}(P_{\tau}, \widehat{P_{\tau}}) = Max \left\{A, B\right\},  where
\label{max}
\end{equation}

\begin{equation}
\small
    A = \frac{1}{\left |P_{\tau} \right|} \sum_{v_{\tau}\in P_{\tau}} \mathop{\min} \limits_{\widehat{v_{\tau}}\in \widehat{P_{\tau}}} \left \|v_{\tau} - \widehat{v_{\tau}}\right \|_{2}
\end{equation}

\begin{equation}
\small
    B = \frac{1}{\left |\widehat{P_{\tau}} \right|} \sum_{\widehat{v_{\tau}} \in \widehat{P_{\tau}}} \mathop{\min} \limits_{v_{\tau}\in P_{\tau}} \left \| \widehat{v_{\tau}} - v_{\tau} \right \|_{2}
\end{equation}
\noindent where the term $ \mathop{\min}\limits_{\widehat{v_{\tau}} \in \widehat{P_{\tau}}} \left \| v_{\tau} - \widehat{v_{\tau}}\right \|_{2}$ enforces that any 3D point $v_{\tau}$ in temporally colorized skeleton cloud $P_{\tau}$ has a matched 3D point $\widehat{v_{\tau}}$ in the repainted point cloud $\widehat{P_{\tau}}$. The term $\mathop{\min} \limits_{v_{\tau} \in P_{\tau}} \left \| \widehat{v_{\tau}} - v_{\tau} \right \|_{2}$ enforces the matching vice versa. The max operation enforces that the distance from $P_{\tau}$ to $\widehat{P_{\tau}}$ and vice versa need to be small concurrently. 

By using the Chamfer distance, the encoder $E_{\tau}(.)$ and decoder $D_{\tau}(.)$ are forced to learn temporal dependency via the proposed temporal repainting scheme that reconstructs temporal-order colors.
Similarly, the encoder $E_{s}(.)$ and decoder $D_{s}(.)$ will learn useful spatial relations during spatial repainting, and the encoder $E_{p}(.)$ and decoder $D_{p}(.)$ are pushed to distinguish the person index and learn interactive information.
It is non-trivial to repaint $P_{r}$ to colorized skeleton clouds. 
To balance the color repainting and unsupervised feature learning, 
we uniformly sample half of points in $P_{r}$ for colorization.
Specifically, in the temporal colorization stream, points corresponding to odd-order frames are colored, while the rest is not.
In the spatial colorization, points with odd-index joints are colored based on joints orders, and even-indexed joints are not colored.

\begin{table}[t]
\caption{Comparisons to state-of-the-art unsupervised skeleton action recognition method on NTU RGB+D dataset. The evaluation setting is as in \cite{kundu2019unsupervised, lin2020ms2l, nie2020view, su2020predict, zheng2018unsupervised}.  (`\textit{TS}':Temporal Stream; `\textit{SS}':Spatial Stream; `\textit{PS}':Person Stream)
}
\vspace{-0.3cm}
\begin{center}
\footnotesize
\begin{tabular}{|l|c|c|}
  \hline 
  \multirow{2}{*}{Method} & \multicolumn{2}{c|}{ \textbf{NTU RGB+D}}\\ 
  \cline{2-3} 
   & C-Subject  & C-View \\ 
  \hline
  \hline
  LongT GAN \cite{zheng2018unsupervised} & 39.1 & 52.1 \\
  M$S^{2}$L \cite{lin2020ms2l}  & 52.6    & --  \\ 
  P\&C FS-AEC \cite{su2020predict} & 50.6 &  76.3\\ 
  P\&C FW-AEC \cite{su2020predict} & 50.7 & 76.1 \\ 
  EnGAN-PoseRNN \cite{kundu2019unsupervised} &68.6 &  77.8 \\
  SeBiReNet \cite{nie2020view}   & -- & 79.7  \\
  \hline
  \hline
  `\textit{TS}' Colorization (Ours)          &  71.6 & 79.9 \\ 
  `\textit{TS} $+$ \textit{SS}' Colorization (Ours)      & 74.6 & 82.6 \\
  `\textit{TS} $+$ \textit{SS} $+$ \textit{PS}' Colorization (Ours)         &  \textbf{75.2} &  \textbf{83.1}\\
  \hline
\end{tabular}
\end{center}
\label{tab:unsupervised result NTU}
\end{table}

\begin{table}[t]
\caption{Comparisons to state-of-the-art unsupervised skeleton action recognition method on NW-UCLA dataset. The evaluation setting is as in \cite{kundu2019unsupervised, lin2020ms2l, nie2020view, su2020predict, zheng2018unsupervised}. 
}
\vspace{-0.3cm}
\begin{center}
\footnotesize
\begin{tabular}{|l|c|}
  \hline 
  Method &  \textbf{NW-UCLA}  \\

  \hline
  \hline
  LongT GAN \cite{zheng2018unsupervised} & 74.3  \\
  M$S^{2}$L \cite{lin2020ms2l}  & 76.8   \\ 
  SeBiReNet  \cite{nie2020view} & 80.3  \\
  P\&C FS-AEC \cite{su2020predict} & 83.8  \\ 
  P\&C FW-AEC \cite{su2020predict} & 84.9 \\ 
  \hline
  \hline
  `\textit{TS}' Colorization (Ours)       & 90.1  \\ 
  `\textit{TS} $+$ \textit{SS}' Colorization (Ours)        & \textbf{91.1}  \\
  \hline
\end{tabular}
\end{center}
\label{tab:unsupervised result NWUCLA}
\end{table}


\begin{table*}[t]
\caption{Comparisons of action recognition results with semi-supervised learning approaches on NTU RGB+D dataset. 
The number in parentheses denotes the number of labeled samples per class.
}
\vspace{-0.3cm}
\begin{center}
\scriptsize
\resizebox{0.95\textwidth}{!}{
\begin{tabular}{l|c|c|c|c|c|c|c|c|c|c}
  \hline 
  \multirow{2}{*}{Method} & \multicolumn{2}{c|}{ \textbf{$1\%$}}&\multicolumn{2}{c|}{ \textbf{$5\%$}}& \multicolumn{2}{c|}{ \textbf{$10\%$}}& \multicolumn{2}{c|}{ \textbf{$20\%$}}& \multicolumn{2}{c}{ \textbf{$40\%$}}\\ 
  \cline{2-11} 
   & CS $(7)$ & CV $(7)$& CS $(33)$& CV $(31)$ & CS $(66)$ & CV $(62)$ & CS $(132)$ & CV $(124)$ & CS $(264)$ & CV $(248)$ \\ 
  \hline
  \hline
  Pseudolabels  \cite{lee2013pseudo}        & --   & -- & 50.9 & 56.3 & 57.2 & 63.1 & 62.4 & 70.4 & 68.0 & 76.8 \\
  VAT    \cite{miyato2018virtual}              & --   & -- & 51.3 & 57.9 & 60.3 & 66.3 & 65.6 & 72.6 & 70.4 & 78.6 \\
  VAT + EntMin          & --   & -- & 51.7 & 58.3 & 61.4 & 67.5 & 65.9 & 73.3 & 70.8 & 78.9 \\
  $ S^{4}L$(Inpainting) \cite{zhai2019s4l} & --   & -- & 48.4 & 55.1 & 58.1 & 63.6 & 63.1 & 71.1 & 68.2 & 76.9 \\
  ASSL \cite{si2020adversarial}& --   & -- & 57.3 & 63.6 & 64.3 & 69.8 & 68.0 & 74.7 & 72.3 & 80.0 \\
  LongT GAN \cite{zheng2018unsupervised} & 35.2 & -- & --   & --   & 62.0 & --   & --   & --   & --   & --\\
  M$S^{2}$L \cite{lin2020ms2l} & 33.1 & -- & --   & --   & 65.2 & --   & --   & --   & --   & --\\
  \hline
  \hline
  `\textit{TS}' Colorization (Ours)          & 42.9   & 46.3 & 60.1 & 63.9 & 66.1 & 73.3 & 72.0 & 77.9 & 75.9 & 82.7 \\ 
  `\textit{TS} $+$ \textit{SS}' Colorization (Ours) & 48.1   & 51.5 & 64.7 & 69.3 & 70.8 & 78.2 & 75.2 & 81.8 & 79.2 & 86.0 \\
  `\textit{TS} $+$ \textit{SS} $+$ \textit{PS}' Colorization (Ours) & \textbf{48.3}  & \textbf{52.5} & \textbf{65.7} & \textbf{70.3} & \textbf{71.7} & \textbf{78.9} & \textbf{76.4} & \textbf{82.7} & \textbf{79.8} & \textbf{86.8} \\
  \hline
\end{tabular}
}
\end{center}
\label{tab:semi-supervised result NTU}
\end{table*}
\begin{table*}[t]
\caption{Comparisons of action recognition results with semi-supervised learning approaches on NW-UCLA dataset. 
$v./c.$ denotes the number of labeled videos per class.
}
\vspace{-0.3cm}
\begin{center}
\scriptsize
\resizebox{0.95\textwidth}{!}{
\begin{tabular}{l|c|c|c|c|c|c}
  \hline 
  Method &  $1\% \; (1 \;v_{.}/c_{.})$ &  $5\% \; (5 \;v_{.}/c_{.})$ &  $10\% \; (10 \;v_{.}/c_{.})$ &  $15\% \; (15 \;v_{.}/c_{.})$ &  $30\% \; (30 \;v_{.}/c_{.})$ &  $40\% \; (40 \;v_{.}/c_{.})$  \\
  \hline
  \hline
  Pseudolabels \cite{lee2013pseudo}          &  --    & 35.6 &  --    & 48.9  & 60,6  & 65.7 \\
  VAT      \cite{miyato2018virtual}               &  --    & 44.8 &  --    & 63.8  & 73.7  & 73.9\\
  VAT + EntMin \cite{grandvalet2005semi}          &  --    & 46.8 &  --    & 66.2  & 75.4  & 75.6\\
  $ S^{4}L$(Inpainting) \cite{zhai2019s4l}  &  --    & 35.3 &  --    & 46.6  & 54.5  & 60.6\\
  ASSL  \cite{si2020adversarial}     &  --    & 52.6 &  --    & 74.8 & 78.0 & 78.4\\
  LongT GAN \cite{zheng2018unsupervised}  &  18.3  & --   & 59.9   & --   & --   & --  \\
  M$S^{2}$L \cite{lin2020ms2l}  &  21.3  & --   & 60.5   & --   & --   & --  \\
  \hline
  \hline
  `\textit{TS}' Colorization (Ours)            & 40.6  &  55.9 &  71.3 &  74.3  & 81.4 & 83.6 \\
  `\textit{TS} $+$ \textit{SS}' Colorization (Ours)  &  \textbf{41.9}  &    \textbf{57.2}    &    \textbf{75.0}    &    \textbf{76.0}    &  \textbf{83.0}    &   \textbf{84.9} \\
  \hline
\end{tabular}
}
\end{center}
\label{tab:semi-supervised result NWUCLA}
\end{table*}

\subsection{Training for Skeleton Action Recognition} \label{training}
After the self-supervised repainting, we obtain three encoders (\textit{i.e.}, $E_{\tau}(.)$. $E_{s}(.)$, and $E_{p}(.)$) that capture meaningful temporal, spatial, interaction features, respectively. With the feature representations from the three encoders, we include a simple linear classifier $f(.)$ on top of the encoder to perform action recognition as in \cite{kundu2019unsupervised, lin2020ms2l, nie2020view, su2020predict, zheng2018unsupervised}. We adopt different settings to train the classifier including unsupervised, semi-supervised, and supervised settings. In the unsupervised setting, the encoder is only trained by the skeleton cloud repainting method, and then we train the linear classifier with the encoder fixed by following previous unsupervised skeleton representation learning works \cite{kundu2019unsupervised, lin2020ms2l, nie2020view, su2020predict, zheng2018unsupervised}. In the semi-supervised and supervised settings, the encoder is first trained with unsupervised representation learning and then fine-tuned with the linear classifier as in \cite{lin2020ms2l}. We use the standard cross-entropy loss as the classification loss $L_{cls}$.
\section{Experiments}
We conducted extensive experiments over two publicly accessible datasets including NTU RGB+D \cite{Shahroudy_2016_CVPR}, and Northwestern-UCLA \cite{wang2014cross}. The experiments aim to evaluate whether our skeleton cloud colorization scheme can learn effective unsupervised feature representations for the task of skeleton action recognition. We therefore evaluate different experimental settings including unsupervised and semi-supervised as well as supervised.

\subsection{Datasets}
{\bf NTU RGB+D \cite{Shahroudy_2016_CVPR}:} NTU RGB+D consists of 56880 skeleton action sequences which is the most widely used dataset for skeleton-based action recognition research. In this dataset, action samples are performed by 40 volunteers and categorized into 60 classes. Each sample contains an action and is guaranteed to have at most two subjects as captured by three Microsoft Kinect v2 cameras from different views. The authors of this dataset recommend two benchmarks: (1) cross-subject (CS) benchmark where training data comes from 20 subjects and testing data comes from the other 20 objects; (2) cross-view (CV) benchmark where training data comes from camera views 2 and 3, and testing data comes from camera view 1.

{\bf Northwestern-UCLA (NW-UCLA) \cite{wang2014cross}:} This dataset is captured by three Kinect v1 cameras and it contains 1494 samples performed by 10 subjects. It contains 10 action classes, and each body has 20 skeleton joints. Following the evaluation protocol in \cite{wang2014cross}, the training set consists of samples from the first two cameras ($V1$, $V2$) and the rest samples from the third camera ($V3$) form the testing set.

\subsection{Implementation Details}
For NTU RGB+D, we adopt the pre-processing in \cite{Shi_2019_CVPR_twostream} and uniformly sampled $T = 40$ frames from each skeleton sequence. The sampled skeleton sequences are constructed to a 2000-points skeleton cloud. For NW-UCLA, we adopt the pre-processing in \cite{su2020predict} and uniformly sampled $T = 50$ frames from the skeleton sequences. The skeleton cloud has 1000 points.

In the unsupervised feature learning phase, we use Adam optimizer and set the initial learning rate at 1e-5 and reduce it to 1e-7 with cosine annealing. The dimension of the encoder output is 1024, and the batch size is 24 for both datasets. The training lasts for 150 epochs. In the classifier training phase, we use SGD optimizer with Nesterov momentum (0.9). We set the initial learning rate at 0.001 and reduce it to 1e-5 with cosine annealing. The batch size for NTU RGB+D and NW-UCLA is 32 and 64 and the training lasts for 100 epochs. We implement our method using PyTorch and all experiments were conducted on Tesla P100 GPUs with CUDA 10.1.

\subsection{Comparison with the state-of-the-art methods}
We conduct extensive experiments under three settings including unsupervised learning, semi-supervised learning and supervised learning. We also study three skeleton cloud colorization configurations: 1) `\textit{T-Stream (TS)}' that uses temporally colorized skeleton cloud as self-supervision; 2) `\textit{S-Stream (SS)}' that uses spatially colorized skeleton cloud as self-supervision; and 3) `\textit{P-Stream (PS)}' that uses person-level colorized cloud as self-supervision.
\begin{table}[t]
\caption{Comparisons to state-of-the-art supervised skeleton action recognition method on NTU RGB+D dataset. 
}
\vspace{-0.3cm}
\begin{center}
\footnotesize
\begin{tabular}{|l|c|c|}
  \hline 
  \multirow{2}{*}{Method} &  \multicolumn{2}{c|}{ \textbf{NTU RGB+D}} \\
  \cline{2-3} & C-Subject & C-View \\
  \hline
  \hline
  \multicolumn{3}{|c|}{ \textbf{Supervised method}} \\
  \hline
  \hline
  ST-LSTM  \cite{liu2016spatio}        & 69.2  & 77.7 \\
  GCA-LSTM \cite{liu2017global}         & 74.4  & 82.8 \\
  ST-GCN \cite{yan2018spatial}           & 81.5  & 88.3 \\
  AS-GCN \cite{li2019actional}           & 86.8  & 94.2 \\
  2s AGC-LSTM \cite{si2019attention}     & 89.2  & 95.0 \\
  4s MS-AAGCN \cite{shi2020skeleton}        & 90.0  & 96.2 \\
  4s Shift-GCN \cite{cheng2020skeleton}  & \textbf{90.7}  &  \textbf{96.5} \\
  \hline
  \hline
  \multicolumn{3}{|c|}{ \textbf{Unsupervised Pretrain}} \\
  \hline
  \hline
  Li \textit{et al.} \cite{li2018unsupervised} & 63.9 & 68.1 \\
  M$S^{2}$L \cite{lin2020ms2l} &    78.6     &   --   \\
  \hline
  `\textit{TS}' Colorization (Ours)      & 84.2   & 93.1 \\ 

  `\textit{TS} $+$ \textit{SS}' Colorization (Ours)  & 86.3   & 94.2 \\
  `\textit{TS} $+$ \textit{SS} $+$ \textit{PS}' Colorization (Ours) & \textbf{88.0}   & \textbf{94.9} \\
  \hline
\end{tabular}
\end{center}
\label{tab:supervised result NTU}
\end{table}

\textbf{Unsupervised Learning}.
In the unsupervised setting, the feature extractor (i.e. the encoder $E(.)$) is trained with our proposed skeleton cloud colorization unsupervised representation learning approach. Then the feature representation is evaluated by the simple linear classifier $f(.)$, which is trained on the top of the frozen encoders $E(.)$. Such experimental setting for unsupervised learning has been widely adopted and practised in prior studies \cite{kundu2019unsupervised, lin2020ms2l, nie2020view, su2020predict, zheng2018unsupervised}. Here for fair comparisons, we use the same setting as these prior works.

We compare our skeleton cloud colorization method with prior unsupervised methods on NTU RGB+D and NW-UCLA datasets as shown in Tables \ref{tab:unsupervised result NTU} and \ref{tab:unsupervised result NWUCLA}. It can be seen that our proposed temporal colorization encoding (i.e. \textit{'TS'} colorization) outperforms prior unsupervised methods on NTU RGB+D dataset, especially under the cross-subject evaluation protocol. Additionally, the proposed `\textit{TS} $+$ \textit{SS}' colorization and `\textit{TS} $+$ \textit{SS} $+$ \textit{PS}' colorization outperform the state-of-the-art obviously on both cross subject and cross view protocols. For NW-UCLA, our method outperforms the state-of-the-art consistently for both configurations `\textit{TS}' and `\textit{TS} $+$ \textit{SS}' as shown in Table \ref{tab:unsupervised result NWUCLA}.

\begin{table}[t]
\caption{Comparisons to state-of-the-art supervised skeleton action recognition method on NW-UCLA dataset. 
}
\vspace{-0.3cm}
\begin{center}
\footnotesize
\begin{tabular}{|l|c|}
  \hline 
  Method &  \textbf{NW-UCLA}  \\

  \hline
  \hline
  \multicolumn{2}{|c|}{ \textbf{Supervised method}} \\
  \hline
  \hline
  Actionlet ensemble \cite{wang2013learning} & 76.0 \\
  HBRNN-L \cite{du2015hierarchical} & 78.5 \\
  Ensemble TS-LSTM \cite{lee2017ensemble} & 89.2 \\
  VA-RNN-Aug  \cite{zhang2019view} & 90.7 \\
  2s AGC-LSTM  \cite{si2019attention}& 93.3  \\
  1s Shift-GCN \cite{cheng2020skeleton}  & 92.5   \\ 
  4s Shift-GCN \cite{cheng2020skeleton}  & \textbf{94.6}  \\
  \hline
  \hline
  \multicolumn{2}{|c|}{ \textbf{Unsupervised Pretrain}} \\
  \hline
  \hline
  Li \textit{et al.} \cite{li2018unsupervised} & 62.5 \\
  M$S^{2}$L \cite{lin2020ms2l} &    86.8   \\
  \hline
  `\textit{TS}' Colorization (Ours)       & 92.7  \\ 
  `\textit{TS} $+$ \textit{SS}' Colorization (Ours)        & \textbf{94.0}  \\
  \hline
\end{tabular}
\end{center}
\label{tab:supervised result NWUCLA}
\end{table}
\begin{table}[t]
\caption{Comparisons of different network configurations' results with unsupervised and supervised setting on NTU RGB+D and NW-UCLA datasets.
}
\vspace{-0.3cm}
\begin{center}
\footnotesize
\begin{tabular}{|l|c|c|c|}
  \hline 
  Dataset &  NTU-CS  &  NTU-CV &  NW-UCLA \\
  \hline
  \hline
  \multicolumn{4}{|c|}{ \textbf{Unsupervised Setting}} \\
  \hline
  \hline
  Baseline-U &    $61.8$     &    $68.4$     &  $78.6$        \\
  \hline
  `\textit{TS} Colorization' &    $71.6$   &    $79.9$     &     $90.1$      \\
  `\textit{SS} Colorization'    &   $68.4$    &    $77.5$    &     $87.0$       \\
  `\textit{PS} Colorization'   &    $64.2$    &      $72.8$ &     --     \\
  \hline
  `\textit{TS} $+$ \textit{SS}' Colorization &  $74.6$  &   $82.6$     & \textbf{91.1} \\
  `\textit{TS} $+$ \textit{PS}' Colorization &  $73.3$   &     $81.4$  &  -- \\
  `\textit{JS} $+$ \textit{PS}' Colorization &   $69.6$   &    $78.6$   &  -- \\
  \hline
  `\textit{TS} $+$ \textit{SS} $+$ \textit{PS}' Colorization &  \textbf{75.2}    &  \textbf{83.1}    & -- \\
  \hline
  \hline
  \multicolumn{4}{|c|}{ \textbf{Supervised Setting}} \\
  \hline
  \hline
  
  Baseline-S &    $76.5$     &   $83.4$      &  $83.8$        \\
  \hline
  `\textit{TS}' Colorization &   $84.2$     &  $93.1$      &     $92.7$      \\
  `\textit{SS}' Colorization    &   $82.3$    &   $91.5$     &     $90.4$       \\
  `\textit{PS}' Colorization   &   $81.1$     &     $90.3$  &     --     \\
  \hline
  `\textit{TS} $+$ \textit{SS}' Colorization &  $86.3$   &  $94.2$     & \textbf{94.0}  \\
  `\textit{TS} $+$ \textit{PS}' Colorization &  $86.4$  &    $94.1$   &  -- \\
  `\textit{SS} $+$ \textit{PS}' Colorization &  $85.0$     &   $93.0$    &  -- \\
  \hline
  `\textit{TS} $+$ \textit{SS} $+$ \textit{PS}' Colorization & \textbf{88.0}    &   \textbf{94.9}   & -- \\
  \hline
\end{tabular}
\end{center}
\label{tab:abla unsup and sup}
\end{table}

\begin{table*}[t]
\caption{Comparisons of different network configurations' results with semi-supervised setting on NW-UCLA dataset. 
}
\begin{center}
\scriptsize
\resizebox{0.95\textwidth}{!}{
\begin{tabular}{l|c|c|c|c|c|c}
  \hline 
  Method &  $1\% \; (1 \;v_{.}/c_{.})$ &  $5\% \; (5 \;v_{.}/c_{.})$ &  $10\% \; (10 \;v_{.}/c_{.})$ &  $15\% \; (15 \;v_{.}/c_{.})$ &  $30\% \; (30 \;v_{.}/c_{.})$ &  $40\% \; (40 \;v_{.}/c_{.})$ \\
  \hline
  \hline
  Baesline-Semi  & 34.3   &  46.4  &  54.9 &  61.8 & 69.1 & 70.2 \\
  \hline
  \hline
  `\textit{TS}' Colorization             & 40.6  &  55.9 &  71.3 &  74.3  & 81.4 & 83.6 \\
  `\textit{SS}' Colorization   & 39.1  &  54.2 &  66.3 &  70.2 & 79.1 & 80.8 \\
  `\textit{TS} $+$ \textit{SS}' Colorization  &  \textbf{41.9}  &    \textbf{57.2}    &    \textbf{75.0}    &    \textbf{76.0}    &  \textbf{83.0}    &   \textbf{84.9} \\
  \hline
\end{tabular}
}
\end{center}
\label{tab:Abla semi NWUCLA}
\end{table*}
\begin{table*}[h]
\caption{Comparisons of different network configurations' results with semi-supervised setting on NTU RGB+D dataset. 
}
\begin{center}
\scriptsize
\resizebox{0.95\textwidth}{!}{
\begin{tabular}{l|c|c|c|c|c|c|c|c|c|c}
  \hline 
  \multirow{2}{*}{Method} & \multicolumn{2}{c|}{ \textbf{$1\%$}}&\multicolumn{2}{c|}{ \textbf{$5\%$}}& \multicolumn{2}{c|}{ \textbf{$10\%$}}& \multicolumn{2}{c|}{ \textbf{$20\%$}}& \multicolumn{2}{c}{ \textbf{$40\%$}}\\ 
  \cline{2-11} 
   & CS $(7)$ & CV $(7)$& CS $(33)$& CV $(31)$ & CS $(66)$ & CV $(62)$ & CS $(132)$ & CV $(124)$ & CS $(264)$ & CV $(248)$ \\ 
  \hline
  \hline
  Baseline-Semi  &  27.1   & 28.1 & 46.0 & 50.6 & 55.1 & 60.7 & 60.9 & 69.1 & 64.2 & 73.7 \\
  \hline
  \hline
   `\textit{TS}' Colorization          & 42.9   & 46.3 & 60.1 & 63.9 & 66.1 & 73.3 & 72.0 & 77.9 & 75.9 & 82.7 \\ 
  `\textit{SS}' Colorization                     & 40.2   & 43.1 & 54.6 & 60.0 & 60.1 & 68.1 & 64.2 & 73.1 & 69.1 & 77.6 \\
  `\textit{PS}' Colorization                     & 37.9   & 40.1 & 51.2 & 56.0 & 56.8 & 63.2 & 61.9 & 70.2 & 65.8 & 74.59 \\
  \hline
  \hline
  `\textit{TS} $+$ \textit{SS}' Colorization  & 48.1   & 51.5 & 64.7 & 69.3 & 70.8 & 78.2 & 75.2 & 81.8 & 79.2 & 86.0 \\
  `\textit{TS} $+$ \textit{PS}' Colorization & 46.9   & 50.5 & 63.9 & 68.1 & 69.8 & 77.0 & 74.9 & 81.3 & 78.3 & 85.4 \\
  `\textit{SS} $+$ \textit{PS}' Colorization & 43.2   & 46.7 & 58.3 & 64.4 & 64.2 & 72.0 & 69.0 & 77.1 & 73.2 & 81.5 \\
  \hline
  \hline
  `\textit{TS} $+$ \textit{SS} $+$ \textit{PS}' Colorization & \textbf{48.3}  & \textbf{52.5} & \textbf{65.7} & \textbf{70.3} & \textbf{71.7} & \textbf{78.9} & \textbf{76.4} & \textbf{82.7} & \textbf{79.8} & \textbf{86.8} \\
  \hline
\end{tabular}
}
\end{center}
\label{tab:abla semi-supervised result NTU}
\end{table*}

\textbf{Semi-supervised Learning}.
We evaluate semi-supervised learning with the same protocol as in \cite{lin2020ms2l, si2020adversarial} for fair comparisons. Under the semi-supervised setting, the encoder $E(.)$ is first pre-trained with colorized skeleton clouds and then jointly trained with the linear classifier $f(.)$ with a small ratio of action annotations. Following \cite{lin2020ms2l, si2020adversarial}, we derive labelled data by uniformly sampling $1\%$, $5\%$, $10\%$, $20\%$, $40\%$ data from the training set of NTU RGB+D dataset, and $1\%$, $5\%$, $10\%$, $15\%$, $30\%$, $40\%$ data from the training set of NW-UCLA dataset, respectively. 

Tables \ref{tab:semi-supervised result NTU} and \ref{tab:semi-supervised result NWUCLA} show experimental results on NTU RGB+D and NW-UCLA datasets, respectively. As Table \ref{tab:semi-supervised result NTU} shows, our method performs better than the state-of-the-art consistently for all three configurations (`\textit{TS}', `\textit{TS} $+$ \textit{SS}', and `\textit{TS} $+$ \textit{SS} $+$ \textit{PS}') on NTU RGB+D. For NW-UCLA dataset, our proposed temporal colorization encoding (`\textit{TS}') performs better than the state-of-the-art as shown in Table \ref{tab:semi-supervised result NWUCLA}. Additionally, our `\textit{TS} $+$ \textit{SS}' colorization scheme outperforms the state-of-the-art by large margins.

\textbf{Supervised Learning}.
Following the supervised evaluation protocol in \cite{lin2020ms2l}, we pre-train the encoder with our unsupervised skeleton colorization method and fine-tune the encoder and classifier by using labeled training data. Tables \ref{tab:supervised result NTU} and \ref{tab:supervised result NWUCLA} show experimental results. We can observe that our method achieves superior performance on NW-UCLA, and it performs much better than the previous ``Unsupervised Pre-trained'' methods \cite{li2018unsupervised, lin2020ms2l} (first train under the unsupervised feature learning and then fine-tune the framework with the labeled data). On the large-scale NTU RGB+D, our method outperforms the ``Unsupervised Pre-trained'' works \cite{li2018unsupervised, lin2020ms2l} by large margins. Though our framework is not designed for supervised setting, its performance is even comparable to state-of-the-art supervised methods.

\subsection{Ablation Study}
\textbf{Effectiveness of Our Skeleton Colorization:} We verify the effectiveness of our skeleton cloud colorization on all three learning settings including unsupervised learning, semi-supervised learning, and fully-supervised learning. We compare our method with three baselines: 1) \textit{Baseline-U}: it only trains the linear classifier and freezes the encoder which is randomly initialized; 2) \textit{Baseline-Semi}: the encoder is initialized with random weight instead of pre-training by our unsupervised representation learning; 3) \textit{Baseline-S}: the same as \textit{Baseline-Semi}. We train the encoder and linear classifier jointly with action labels. The input for these three baselines is the raw skeleton cloud without color label information.

Tables \ref{tab:abla unsup and sup}, \ref{tab:Abla semi NWUCLA} and \ref{tab:abla semi-supervised result NTU} show experimental results. It can be seen that all the three colorization strategies (i.e. temporal-level, spatial-level, and person-level) achieve significant performance improvement as compared with the baseline, demonstrating the effectiveness of our proposed colorization technique. Though the person-level colorization stream does not perform as well as the other two streams on the NTU RGB+D, it improves the overall performance while collaboration with the other two.

\textbf{Effectiveness of Colorization Ratio:} As mentioned in Section \ref{Ratio of mask}, we need to balance between the repainting and the spatial-temporal ordered feature learning. We observe that the unsupervised performance improves from $65.7\%$ to $71.6\%$ on the temporal stream on the NTU RGB+D (Cross-subject setting) when 50\% points with the color information is provided, demonstrating the effectiveness of our proposed colorization scheme. Detailed comparisons can be found in the supplementary materials.


\section{Conclusion}
\vspace{-0.1cm}
In this paper, we address unsupervised representation learning in skeleton action recognition, and design a novel skeleton cloud colorization method that is capable of learning skeleton representations from unlabeled data.
We obtain colored  skeleton  cloud  representations by stacking skeleton sequences to 3D  skeleton cloud and colorizing each point according to its temporal and spatial orders in the skeleton  sequences. Additionally, spatio-temporal features are learned effectively from the corresponding joints’ colors from unlabeled data. 
The experiments demonstrate that our proposed method achieves superior unsupervised action recognition performance. 

\section*{Acknowledgement}
This work was done at Rapid-Rich Object Search (ROSE) Lab, Nanyang Technological University. This research is supported in part by the NTU-PKU Joint Research Institute (a collaboration between the Nanyang Technological University and Peking University that is sponsored by a donation from the Ng Teng Fong Charitable Foundation), the Science and Technology Foundation of Guangzhou Huangpu Development District under Grant 2019GH16, and China-Singapore International Joint Research Institute under Grant 206-A018001. This research work is also partially supported by the SUTD project PIE-SGP-Al-2020-02.

{\small
\bibliographystyle{ieee_fullname}
\bibliography{egbib}
}

\end{document}